\documentclass[conference]{IEEEtran}
\IEEEoverridecommandlockouts

\usepackage{cite}
\usepackage{amsmath,amssymb,amsfonts}
\usepackage{algorithmic}
\usepackage{graphicx}
\usepackage{textcomp}
\usepackage{xcolor}
\usepackage{booktabs}
\usepackage{multirow}
\usepackage{float}
\usepackage[caption=false,font=footnotesize]{subfig}
\usepackage{pifont}

\graphicspath{{Figure/}}

\def\BibTeX{{\rm B\kern-.05em{\sc i\kern-.025em b}\kern-.08em
    T\kern-.1667em\lower.7ex\hbox{E}\kern-.125emX}}

\makeatletter
\newcommand{\twocolumnfootnotefullwidth}[1]{%
  \begingroup
  \renewcommand{\thefootnote}{}
  \footnotetext{%
    \noindent\hspace*{-1em}\rule{0.3\linewidth}{0.4pt}\\[0.2em] 
    \vspace*{-1em}
    \noindent\footnotesize #1
  }
  \endgroup
}
\makeatother

\begin{document}

\title{NEMESIS: Noise-suppressed Efficient MAE with Enhanced
Superpatch Integration Strategy}

\author{
\makebox[\textwidth][c]{%
\resizebox{\textwidth}{!}{%
\begin{tabular}{cccc}
Kyeonghun Kim & Hyeonseok Jung & Youngung Han & Hyunsu Go \\
\textit{OUTTA} & \textit{Chung-Ang University} & \textit{Seoul National University} & \textit{Seoul National University} \\
kyeonghun.kim@outta.ai & mjmk0820@cau.ac.kr & yuhan@snu.ac.kr & hsmail02@snu.ac.kr
\end{tabular}%
}%
}%
\\[0.6ex]

\makebox[\textwidth][c]{%
\resizebox{\textwidth}{!}{%
\begin{tabular}{cccc}
Eunseob Choi & Seongbin Park & Junsu Lim & Jiwon Yang \\
\textit{GIST} & \textit{Seoul National University} & \textit{Sangmyung University} & \textit{Seoul National University} \\
eunseobchoi@gm.gist.ac.kr & tjdqls@snu.ac.kr & 202115042@sangmyung.ac.kr & jwyang29@snu.ac.kr
\end{tabular}%
}%
}%
\\[0.6ex]

\makebox[\textwidth][c]{%
\resizebox{\textwidth}{!}{%
\begin{tabular}{cccc}
Sumin Lee & Insung Hwang & Ken Ying-Kai Liao & Nam-Joon Kim\textsuperscript{\dag} \\
\textit{Seoul National University} & \textit{Seoul National University} & \textit{NVIDIA} & \textit{Seoul National University} \\
cirtuare@snu.ac.kr & insung0608@snu.ac.kr & kenyingkail@nvidia.com & knj01@snu.ac.kr
\end{tabular}%
}%
}%
}

\maketitle

\twocolumnfootnotefullwidth{\quad {\dag} Corresponding author}

\begin{abstract}
Volumetric CT imaging is essential for clinical diagnosis, yet annotating 3D volumes is expensive and time-consuming, motivating self-supervised learning (SSL) from unlabeled data. However, applying SSL to 3D CT remains challenging due to the high memory cost of full-volume transformers and the anisotropic spatial structure of CT data, which is not well captured by conventional masking strategies. We propose NEMESIS, a masked autoencoder (MAE) framework that operates on local $128^{3}$ superpatches, enabling memory-efficient training while preserving anatomical detail. NEMESIS introduces three key components: (i) noise-enhanced reconstruction as a pretext task, (ii) Masked Anatomical Transformer Blocks (MATB) that perform dual-masking through parallel plane-wise and axis-wise token removal, and (iii) NEMESIS Tokens (NT) for cross-scale context aggregation. On the BTCV multi-organ classification benchmark, NEMESIS with a frozen backbone and a linear classifier achieves a mean AUROC of 0.9633, surpassing fully fine-tuned SuPreM (0.9493) and VoCo (0.9387). Under a low-label regime with only 10\% of available annotations, it retains an AUROC of 0.9075, demonstrating strong label efficiency. Furthermore, the superpatch-based design reduces computational cost to 31.0~GFLOPs per forward pass, compared to 985.8~GFLOPs for the full-volume baseline, providing a scalable and robust foundation for 3D medical imaging.
\end{abstract}

\section{Introduction}

Volumetric Computed Tomography (CT) imaging is essential for clinical diagnosis, yet training deep models for CT analysis is limited by the scarcity of expert annotations, which are costly and time-consuming. Self-supervised learning (SSL) offers a promising alternative by leveraging unlabeled data~\cite{chen2021simclr,grill2020byol,he2022mae}, but extending SSL to 3D CT remains challenging.

Two key challenges arise in 3D CT representation learning. First, full-volume Vision Transformers (ViT) incur prohibitive memory costs, making high-resolution processing impractical on standard hardware~\cite{dosovitskiy2021vit,vaswani2017attention}. For instance, a $512 \times 512 \times 400$ volume yields over 32k tokens with $16^3$ patching, exceeding the quadratic complexity of self-attention. Second, CT volumes are inherently anisotropic, with distinct structural characteristics across axial, coronal, and sagittal planes, which are often overlooked by existing masking strategies.

Recent methods such as SuPreM and VoCo improve 3D medical SSL via auxiliary supervision or contrastive objectives, but they do not jointly address memory efficiency and structural anisotropy~\cite{suprem,voco}. To tackle these limitations, we propose \textbf{NEMESIS}, a memory-efficient masked autoencoder framework for 3D CT representation learning.

NEMESIS operates on randomly sampled $128^3$ superpatches to reduce computational overhead while preserving local anatomical structure. We introduce Masked Anatomical Transformer Blocks (MATB), which apply dual-masking across plane-wise and axis-wise streams to capture anisotropic dependencies. In addition, NEMESIS tokens (NT) aggregate coarse-grained global context to guide fine-grained reconstruction.

The primary contributions of this work are as follows:

\begin{itemize}
    \item \textbf{Superpatch Pretraining:} A memory-efficient MAE framework that reduces GFLOPs by $32\times$ compared to full-volume encoders while improving reconstruction quality.
    \item \textbf{MATB Dual-Masking:} A transformer block that captures CT-specific structural inductive biases via parallel masking streams.
    \item \textbf{Superior Downstream Performance:} A frozen NEMESIS backbone with a linear classifier achieves 0.9633 AUROC on BTCV, outperforming fully fine-tuned SOTA models such as SuPreM (0.9493) and VoCo (0.9387).
    \item \textbf{Label Efficiency:} With only 10\% labeled data, NEMESIS approaches the performance of models trained on full supervision.
\end{itemize}

\begin{figure*}[t]
    \centering
    \includegraphics[width=0.9\textwidth]{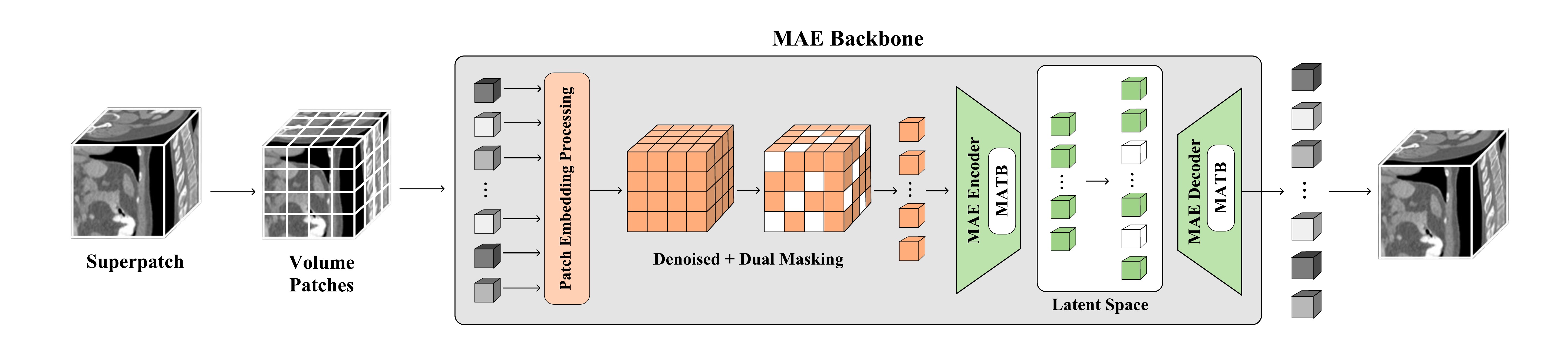}
    \caption{The overall architecture of the NEMESIS backbone. The input superpatch is divided into volume patches and processed through the patch embedding module. The framework utilizes dual-masking and MATB-based encoders and decoders to learn high-level anatomical representations in the latent space.}
    \label{fig:patch_embedding_large}
\end{figure*}
\section{Related Works}

\subsection{SSL for 3D Medical Imaging}
To adapt self-supervised learning (SSL) to medical imaging, early studies explored 3D extensions of pretext tasks such as rotation prediction and Jigsaw puzzles. Models such as Med3D and Models Genesis further investigated generic 3D representation learning through multi-task supervision and restoration-based objectives~\cite{med3d,modelsgenesis}. More recently, SuPreM introduced supervised auxiliary signals to guide pretraining, while VoCo proposed a volume contrastive framework to capture both global and local contextual dependencies~\cite{suprem,voco}. MAESIL further explored superpatch-based masked autoencoding for 3D CT representation learning, showing that chunk-based processing can improve reconstruction quality while preserving 3D contextual information~\cite{maesil}. However, these methods still rely on either partial supervision or substantial memory overhead when processing large-scale 3D volumes with Transformer-based architectures.

\subsection{Efficiency and Anisotropy in 3D Transformers}
Vision Transformers (ViT) have shown strong ability in modeling long-range dependencies, but their quadratic complexity with respect to the number of tokens makes full-volume 3D processing infeasible on standard hardware~\cite{vaswani2017attention,dosovitskiy2021vit}. In addition, CT scans are inherently anisotropic, with spatial resolution and anatomical continuity differing across axial, coronal, and sagittal planes. Most existing 3D SSL frameworks either assume isotropic structure or process 3D volumes as independent 2D slices, failing to exploit CT-specific structural inductive biases. NEMESIS addresses these limitations through a memory-efficient superpatch-based MAE framework and Masked Anatomical Transformer Blocks (MATB) that explicitly model CT anisotropy via dual-masking streams.
\section{Methodology}

\subsection{Superpatch Pretraining Pipeline}

\begin{figure}[H]
    \centering
    \includegraphics[width=\columnwidth]{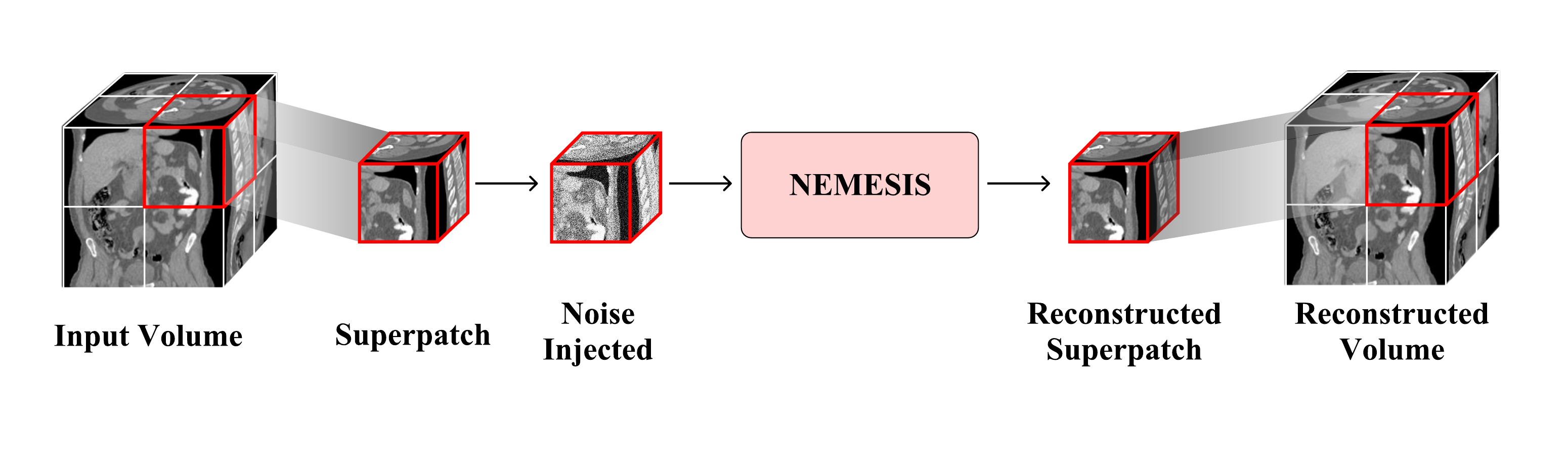}    
    \caption{Overview of the superpatch-based pretraining pipeline. NEMESIS randomly crops $128^3$ superpatches from the original 3D CT volume to ensure memory efficiency and robust feature extraction.}
    \label{fig:pipeline_overview}
\end{figure}

To overcome the memory constraints of 3D ViTs, NEMESIS adopts a superpatch-based approach. As illustrated in Fig.~\ref{fig:pipeline_overview}, a 3D input volume $V \in \mathbb{R}^{H \times W \times D}$ is partitioned into $N$ fixed-size, non-overlapping 3D chunks termed Superpatches $S_i$:
\begin{equation}
    V = \bigcup_{i=1}^{N} S_i, \quad S_i \in \mathbb{R}^{128 \times 128 \times 128}.
\end{equation}

At each training iteration, we randomly sample a superpatch $S$ from the input volume and corrupt it with Gaussian noise to obtain $\tilde{S}$. The model is trained to reconstruct the original clean superpatch $S$ from the visible tokens. During inference, the volume is divided into non-overlapping superpatches that are processed independently and stitched together, enabling memory-efficient full-volume reconstruction.

\subsection{Patch Embedding with NEMESIS Tokens}

The input superpatch is divided into standardized $16^3$ patches $x_{patch}$ and projected into an embedding space. We introduce a 3D Adaptive Patch Embedding module utilizing parallel pathways. 

\begin{figure}[H]
    \centering
    \includegraphics[width=\columnwidth]{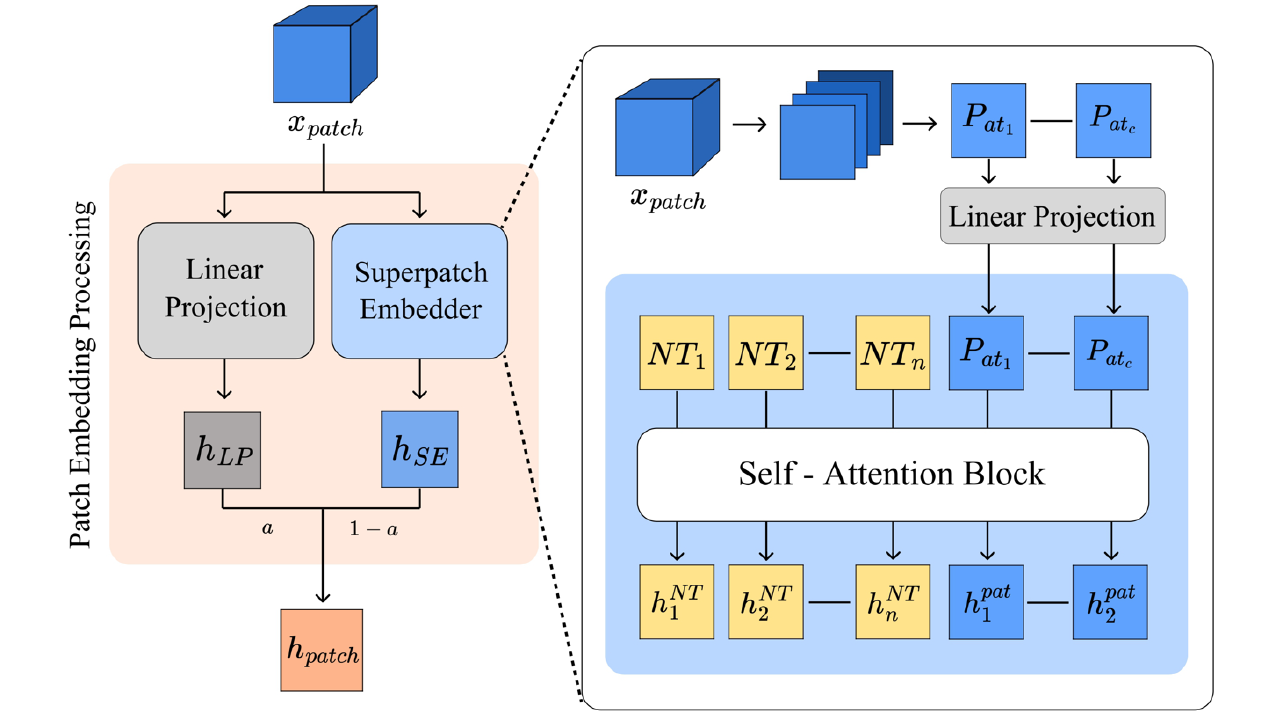}    
    \caption{Detailed architecture of the 3D Adaptive Patch Embedding module, integrating Linear Projection (LP) and Superpatch-Embedder (SE) pathways.}
    \label{fig:patch_embedding}
\end{figure}

A direct Linear Projection pathway generates basic features:
\begin{equation}
    h_{LP} = W_{LP} \cdot x_{patch} + b_{LP}.
\end{equation}
Concurrently, a Superpatch-Embedder pathway utilizes a Self-Attention Block (SAB) to generate context-aware embeddings $h_{SE}$. Critically, the SAB incorporates learnable \textbf{NEMESIS Tokens} (NT) as auxiliary contextual anchors:
\begin{equation}
    h_{SE} = SAB([x_{patch}; NT]).
\end{equation}

The final embedding $h_{patch}$ is fused via a learnable adaptive gate $\alpha$:
\begin{equation}
    h_{patch} = \alpha \cdot h_{LP} + (1-\alpha) \cdot h_{SE}.
\end{equation}

\subsection{Masked Anatomical Transformer Block (MATB)}
To address the inherent anisotropy of CT data, we propose the Masked Anatomical Transformer Block (MATB). 

\begin{figure}[H]
    \centering
    \includegraphics[width=\columnwidth]{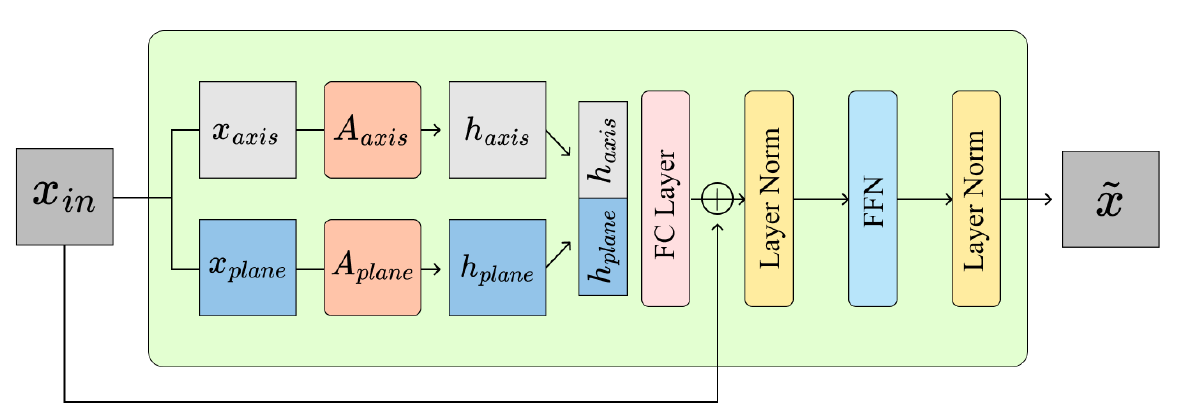}    
    \caption{Detailed architecture of the MATB, featuring parallel axis-wise and plane-wise attention streams to capture comprehensive 3D context.}
    \label{fig:matb_detail}
\end{figure}

As shown in Fig. 4, MATB processes the input through parallel attention streams:
\begin{itemize}
    \item \textbf{Axis-wise stream:} Captures intra-patch context ($h_{axis}$).
    \item \textbf{Plane-wise stream:} Learns inter-patch relationships ($h_{plane}$).
\end{itemize}

The information from these streams is merged via a Fully Connected (FC) layer:
\begin{equation}
    h_{fused} = FC(Concat(h_{axis}, h_{plane})).
\end{equation}
This structure enables the model to effectively learn structural inductive biases tailored for 3D medical imaging.

\subsection{Training Objective}
NEMESIS is trained to minimize the Mean Squared Error (MSE) between the original patches $x_{patch}$ and the reconstructed patches $\hat{x}_{patch}$, calculated only over the masked regions $M$:
\begin{equation}
    \mathcal{L} = \frac{1}{|M|}\sum_{j \in M} \|x_{patch}(j) - \hat{x}_{patch}(j)\|_2^2.
\end{equation}
By minimizing this objective, the encoder extracts highly generalizable anatomical features for downstream tasks.
\section{Experiments}

\subsection{Experimental Setup}
\textbf{Pretraining Data:} For self-supervised pretraining, we use TotalSegmentator~\cite{totalsegmentator} and mixed public CT datasets, including HNSCC, FLARE23, TCIA-COVID19, LUNA16, and BTCV~\cite{hnscc,flare23,tciacovid19,luna16,btcv}, without labels for reconstruction.

\textbf{Dataset and Task:} We evaluate NEMESIS on BTCV~\cite{btcv}, which consists of 30 abdominal CT scans with annotations for 8 target organs: aorta, gallbladder, spleen, left kidney, right kidney, liver, stomach, and pancreas. Our downstream task is superpatch-level multi-label organ-presence classification, where the model predicts whether a superpatch contains each of these organs. We split the dataset into 18 scans for training and 12 for testing, and define superpatch-level labels using a voxel threshold of 100.

\begin{table}[H]
\centering
\caption{Summary of datasets and tasks.}
\label{tab:datasets}
\setlength{\tabcolsep}{3pt}
\renewcommand{\arraystretch}{1.15}
\begin{tabular}{p{2.35cm} c c p{2.1cm} c}
\toprule
\textbf{Dataset} & \textbf{Modality} & \textbf{\# Scans} & \textbf{Task} & \textbf{\# Classes} \\
\midrule
\multicolumn{5}{l}{\textit{Pretraining (Unlabeled)}} \\
TotalSegmentator         & CT & 28  & MAE                  & N/A \\
Mixed public CT datasets & CT & 140 & MAE                  & N/A \\
\midrule
\multicolumn{5}{l}{\textit{Downstream (Labeled)}} \\
BTCV                     & CT & 30  & Organ-presence cls.  & 8 \\
\bottomrule
\end{tabular}
\end{table}

\textbf{Baselines and Protocol:} We compare our model against a supervised ResNet3D-50 baseline~\cite{resnet3d} and state-of-the-art 3D SSL methods, including SuPreM~\cite{suprem} and VoCo~\cite{voco}. While the baseline models are fully fine-tuned, NEMESIS is primarily evaluated with a \textbf{frozen backbone and a linear classifier} to assess the quality of its learned representations. Performance is measured using mean AUROC and macro F1.

\subsection{Main Results}
Table \ref{tab:main_results} presents the quantitative comparison on BTCV organ-presence classification. Notably, \textbf{NEMESIS with a frozen backbone and a linear classifier} achieves a mean AUROC of 0.9633 and an F1 score of 0.753, surpassing the \textbf{fully fine-tuned} results of prior SOTA 3D SSL models, including SuPreM (0.9493 AUROC) and VoCo (0.9387 AUROC).

\begin{table}[htbp]
\centering
\caption{Comparison of NEMESIS with baseline methods on BTCV organ-presence classification.}
\label{tab:main_results}
\setlength{\tabcolsep}{3pt}
\renewcommand{\arraystretch}{1.15}
\begin{tabular}{p{2.0cm} p{3.35cm} c c}
\toprule
\textbf{Method} & \textbf{Protocol} & \textbf{AUROC} & \textbf{F1} \\
\midrule
ResNet3D-50             & Fine-tune                           & 0.9720 & 0.782 \\
SuPreM                  & Fine-tune                           & 0.9493 & 0.678 \\
VoCo                    & Fine-tune                           & 0.9387 & 0.701 \\
Random ViT              & Fine-tune                           & 0.5602 & 0.000$^\dagger$ \\
\midrule
\textbf{NEMESIS (Ours)} & \textbf{Frozen backbone + linear classifier} & \textbf{0.9633} & \textbf{0.753} \\
\textbf{NEMESIS (Ours)} & \textbf{Fine-tune}                  & \textbf{0.9702} & \textbf{0.791} \\
\bottomrule
\multicolumn{4}{p{7.5cm}}{\footnotesize $^\dagger$F1 is reported at a threshold of 0.5.}
\end{tabular}
\end{table}

This performance gap is particularly significant for two reasons. First, it demonstrates that our generative pre-training strategy extracts more informative anatomical features than methods relying on supervised auxiliary signals or contrastive objectives, even without task-specific adaptation of the backbone. Second, the substantial improvement in F1 score (0.753 vs. SuPreM's 0.678) suggests that NEMESIS captures organ boundaries more precisely, which we attribute to the structural inductive bias provided by our dual-masking MATB.

Furthermore, when fully fine-tuned, NEMESIS achieves an F1 score of 0.791, outperforming the strong supervised ResNet3D-50 baseline (0.782). This highlights the capability of our framework to serve as a superior initialization for 3D medical imaging tasks. Overall, the results confirm that by addressing CT anisotropy via MATB, NEMESIS learns representations that are both robust and readily separable for downstream organ-presence classification. The Random ViT baseline shows near-random ranking performance and yields zero F1 at the standard threshold of 0.5.

\subsection{Label Efficiency}
NEMESIS demonstrates strong label efficiency in low-data regimes, which is critical for medical imaging tasks where expert annotations are scarce and costly. 

\begin{table}[htbp]
\centering
\caption{Performance across different label ratios (AUROC).}
\label{tab:label_efficiency}
\addtolength{\tabcolsep}{-2pt}
\renewcommand{\arraystretch}{1.3}
\begin{tabular}{lcccc}
\toprule
\textbf{Method} & \textbf{10\%} & \textbf{25\%} & \textbf{50\%} & \textbf{100\%} \\
\midrule
NEMESIS (Linear) & 0.9075 & 0.9347 & 0.9570 & 0.9633 \\
ResNet3D-50 & 0.9191 & 0.9473 & 0.9612 & 0.9720 \\
SuPreM & 0.8826 & 0.9252 & 0.9409 & 0.9493 \\
VoCo & 0.8664 & 0.8940 & 0.9265 & 0.9387 \\
\bottomrule
\end{tabular}
\end{table}

As shown in Table~\ref{tab:label_efficiency}, with only \textbf{10\% of labels}, NEMESIS with a frozen backbone and a linear classifier achieves an AUROC of 0.9075, indicating that the proposed superpatch-based pretraining learns discriminative anatomical representations even under limited supervision. As the label ratio increases, NEMESIS maintains stable performance gains, highlighting its potential to reduce annotation burden in practical clinical applications.

\subsection{Ablation Studies}
\textbf{Pretraining Strategy:} We evaluate the impact of our dual-masking strategy in Table~\ref{tab:ablation_strategy}. All variants are trained with the same masking ratio of 0.75. Plane-only masking achieves higher reconstruction quality than axis-only masking (PSNR 27.75 vs. 24.87), highlighting the importance of slice-wise structural information in CT volumes.

\begin{table}[htbp]
\centering
\caption{Reconstruction quality under different masking strategies at a masking ratio of 0.75.}
\label{tab:ablation_strategy}
\renewcommand{\arraystretch}{1.3}
\begin{tabular}{lccc}
\toprule
\textbf{Strategy} & \textbf{PSNR (dB) $\uparrow$} & \textbf{SSIM $\uparrow$} & \textbf{LPIPS $\downarrow$} \\
\midrule
Axis-only masking  & 24.87 & 0.294 & 0.392 \\
Plane-only masking & 27.75 & 0.529 & 0.164 \\
\bottomrule
\end{tabular}
\end{table}

\textbf{Hyperparameters:} Table~\ref{tab:hyperparameter} shows that a masking ratio of 0.75 with an embedding dimension of 768 yields the best performance (AUROC 0.9643, F1 0.781).

\begin{table}[htbp]
\centering
\caption{Hyperparameter sensitivity analysis.}
\label{tab:hyperparameter}
\renewcommand{\arraystretch}{1.3}
\begin{tabular}{cccc}
\toprule
\textbf{Embedding dim.} & \textbf{Mask ratio} & \textbf{AUROC} & \textbf{F1} \\
\midrule
384 & 0.50 & 0.9660 & 0.755 \\
576 & 0.50 & 0.9605 & 0.760 \\
768 & 0.50 & 0.9633 & 0.753 \\
\textbf{768} & \textbf{0.75} & \textbf{0.9643} & \textbf{0.781} \\
\bottomrule
\end{tabular}
\end{table}

\subsection{Computational Efficiency}
\begin{figure}[H]
    \centering
    \includegraphics[width=\columnwidth]{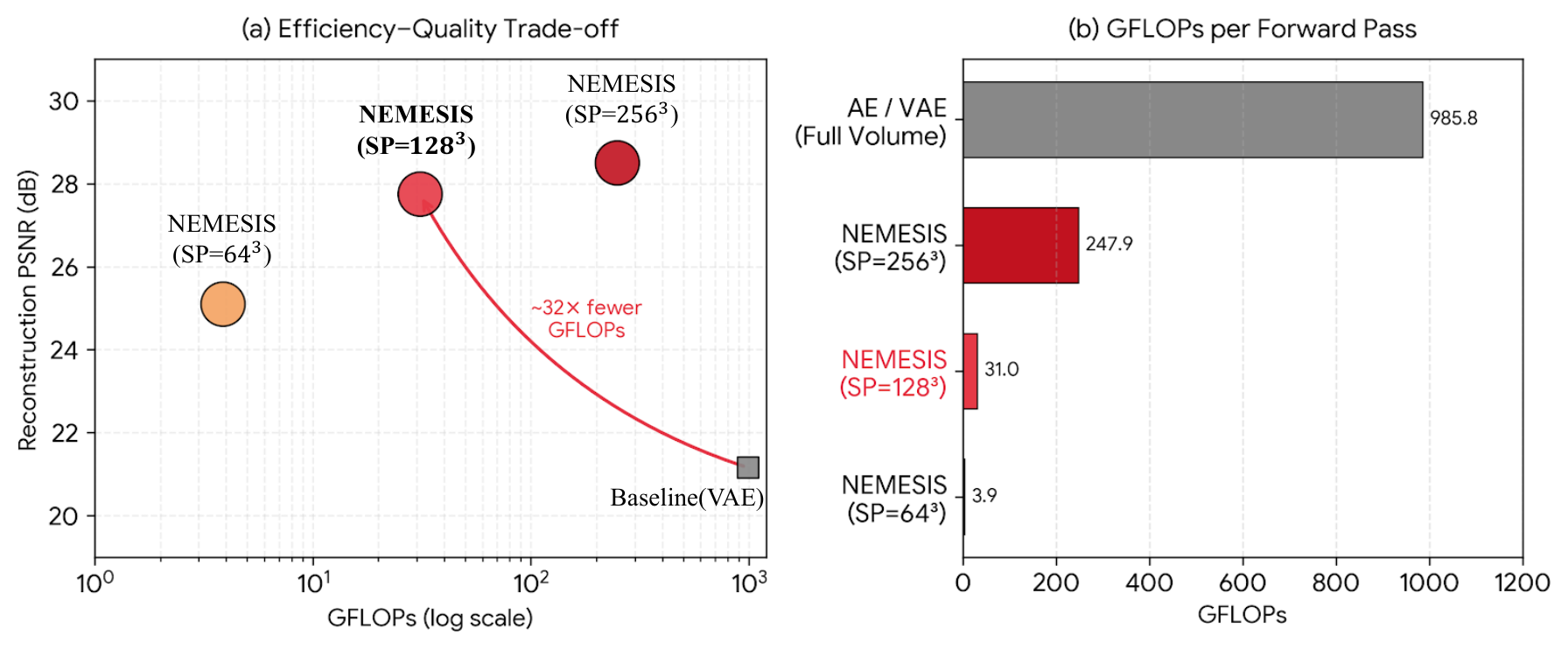}    
    \caption{Computational efficiency analysis. (a) Efficiency-quality trade-off. (b) GFLOPs per forward pass. NEMESIS ($SP=128^3$) achieves a $32\times$ reduction in GFLOPs compared to the VAE baseline while maintaining superior reconstruction quality.}
    \label{fig:Computational_Efficiency}
\end{figure}

NEMESIS substantially alleviates the computational overhead of 3D CT representation learning. As illustrated in Fig.~\ref{fig:Computational_Efficiency}(a), our framework achieves a better trade-off between reconstruction fidelity and computational cost than conventional full-volume baselines. In particular, the NEMESIS ($SP=128^3$) configuration requires only 31.0 GFLOPs, corresponding to a $32\times$ reduction compared with the full-volume baseline (985.8 GFLOPs). Despite this substantial gain in efficiency, NEMESIS maintains a high reconstruction PSNR of 27.75 dB, demonstrating that superpatch-based pretraining enables efficient and effective SSL for 3D CT volumes.

\section{Conclusion}

In this paper, we presented NEMESIS, a memory-efficient self-supervised pretraining framework designed for 3D medical imaging. By leveraging superpatch-based training and the Masked Anatomical Transformer Block (MATB), our approach effectively captures the anisotropic structural biases inherent in CT volumes. Experimental results on the BTCV benchmark demonstrate that a frozen NEMESIS linear probe can outperform state-of-the-art models, such as SuPreM and VoCo, even when they are fully fine-tuned. Furthermore, NEMESIS exhibits exceptional label efficiency, nearly matching the performance of fully supervised models with only 10\% of available annotations, while simultaneously providing a 32$\times$ reduction in computational complexity compared to standard full-volume autoencoders. Future research will focus on extending this framework to more complex segmentation tasks, including the Synapse multi-organ dataset, and investigating the interpretability of the attention mechanisms within the MATB.

\end{document}